\documentclass[sigconf]{acmart}
\renewcommand\footnotetextcopyrightpermission[1]{} 
\AtBeginDocument{%
  }

\copyrightyear{2025}
\acmYear{2025}
\acmConference[arXiv]{}{2025}{}
\acmBooktitle{arXiv}

\begin{document}

\title{AthleticsPose: Authentic Sports Motion Dataset on Athletic Field and Evaluation of Monocular 3D Pose Estimation Ability}

\author{Tomohiro Suzuki}
\affiliation{%
  \institution{Nagoya University}
  \city{Nagoya}
  \country{Japan}}
\email{suzuki.tomohiro@g.sp.m.is.nagoya-u.ac.jp}

\author{Ryota Tanaka}
\affiliation{%
  \institution{Nagoya University}
  \city{Nagoya}
  \country{Japan}}
\email{tanaka.ryota@g.sp.m.is.nagoya-u.ac.jp}

\author{Calvin Yeung}
\affiliation{%
  \institution{Nagoya University}
  \city{Nagoya}
  \country{Japan}}
\email{yeung.chikwong@g.sp.m.is.nagoya-u.ac.jp}

\author{Keisuke Fujii}
\affiliation{%
  \institution{Nagoya University}
  \city{Nagoya}
  \country{Japan}}
\email{fujii@i.nagoya-u.ac.jp}
\renewcommand{\shortauthors}{Suzuki et al.}

\begin{abstract}
Monocular 3D pose estimation is a promising, flexible alternative to costly motion capture systems for sports analysis.
However, its practical application is hindered by two factors: a lack of realistic sports datasets and unclear reliability for sports tasks.
To address these challenges, we introduce the AthleticsPose dataset, a new public dataset featuring ``real'' motions captured from 23 athletes performing various athletics events on an athletic field.
Using this dataset, we trained a representative 3D pose estimation model and performed a comprehensive evaluation.
Our results show that the model trained on AthleticsPose significantly outperforms a baseline model trained on an imitated sports motion dataset, reducing MPJPE by approximately $75\%$.
These results show the importance of training on authentic sports motion data, as models based on imitated motions do not effectively transfer to real-world motions.
Further analysis reveals that estimation accuracy is sensitive to camera view and subject scale.
In case studies of kinematic indicators, the model demonstrated the potential to capture individual differences in knee angles but struggled with higher-speed metrics, such as knee-drive velocity, due to prediction biases.
This work provides the research community with a valuable dataset and clarifies the potential and practical limitations of using monocular 3D pose estimation for sports motion analysis.
Our dataset, code, and checkpoints are available at \url{https://github.com/SZucchini/AthleticsPose}.
\end{abstract}





\maketitle

\section{Introduction}

Objective evaluation of human movement is a basic element of sports science. It provides valuable insights for improving performance, preventing injuries, and supporting decisions \cite{yeung2024autosoccerpose, suzuki2024automatic, tanaka2023automatic_mmsports, tanaka2023automatic_gcce, suzuki2022automatic, piergiovanni2019early_mlbinjury, uhlrich2023opencap}. The gold standard for motion analysis is marker-based motion capture, which provides highly precise 3D joint position. However, the practical application of these systems in daily training is often limited due to their high cost, time-consuming setup procedures, and the constraints of indoor laboratory environments.

In response to these limitations, monocular 2D \& 3D pose estimation \cite{motionagformer2024, liu2025tcpformer, motionbert2022, strided_transformer, videopose3d, sun2019deep_hrnet, 8765346openpose} has been developed as a flexible and accessible alternative. This technology only requires a single video camera, offering portability and cost-effectiveness that make it suitable for field analysis. Despite its potential, its practical utility in sports remains unproven. It is unclear whether current models can accurately detect the critical variations in sports motions that distinguish different levels of athletic performance. This uncertainty comes from two primary challenges.
The first challenge is the lack of appropriate datasets. Most large-scale datasets focus on general daily activities \cite{ionescu2013h36m, mehta2017mpiinf, 3dpw, Humaneva, cocodataset}, which do not reflect the high-speed, dynamic features of sports motions. Although some sports motion datasets exist \cite{AIST++, ASPset, ingwersen2023sportspose, MMSports2024_tanaka_fsjump3d, yeung2025athletepose3d, jiang2024worldpose}, most consist of imitative actions that differ from actual sports motions. The second challenge is the absence of comprehensive evaluation protocols. While most studies focus on positional error metrics, sports applications require the validation of motion metrics, such as joint angles and velocities, derived from estimated poses.

\begin{figure*}[t]
  \centering
  \includegraphics[width=\linewidth]{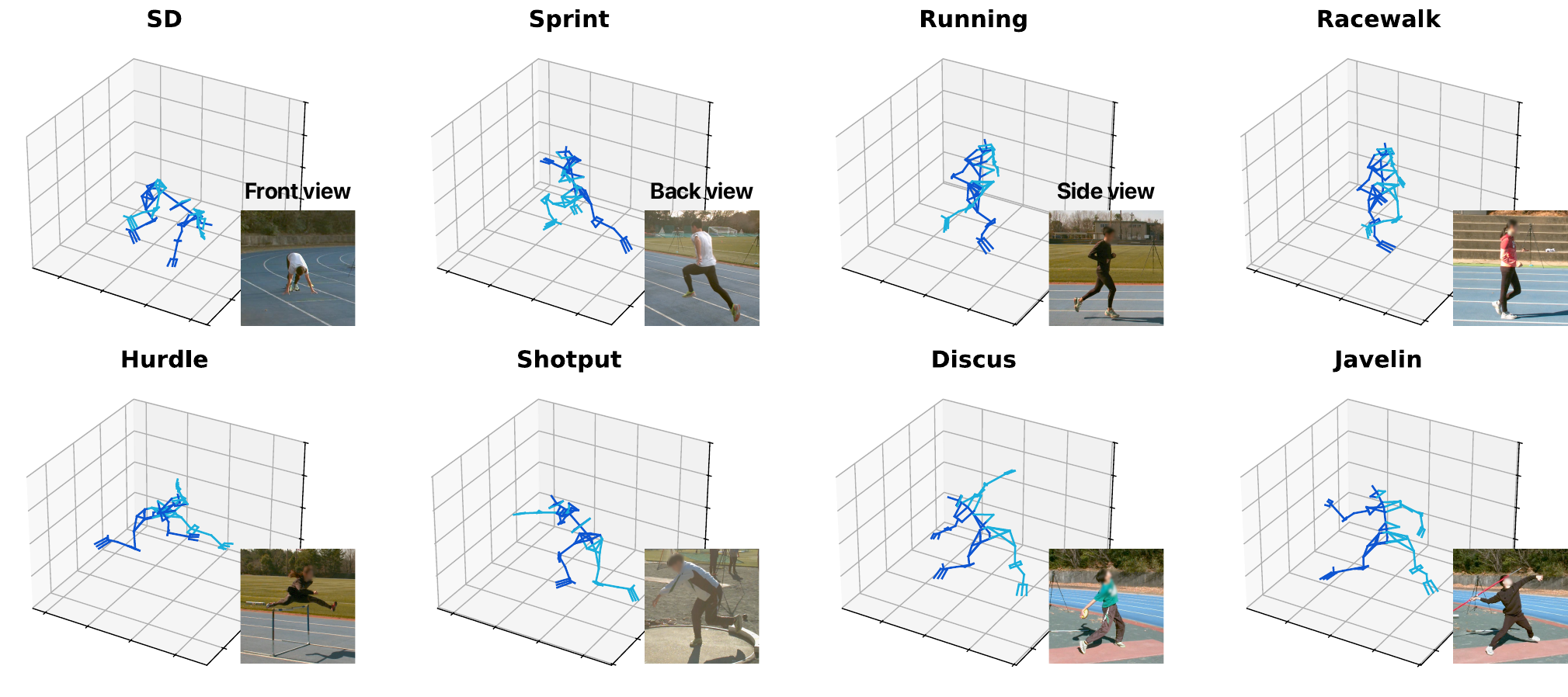}
  \caption{An overview of AthleticsPose dataset, containing authentic motions on an athletic field, with example camera views.}
  \label{fig:dataset_overview}
\end{figure*}

This paper addresses these gaps. First, we introduce AthleticsPose, a new public dataset featuring authentic athletics motions. These motions were captured using eight synchronized cameras with 23 athletes in a realistic outdoor setting to ensure that the data reflects real performance conditions. Second, we evaluate a representative 3D pose estimation model using this dataset. Our analysis exceeds standard error metrics to evaluate the model's ability to offer practical, sport-specific kinematic data.
The main contributions of our work are as follows:
\begin{itemize}
    \item We developed AthleticsPose, the first authentic athletics motion dataset featuring motions from competitive athletics on an athletic field.
    \item We evaluated a representative pose estimation model to assess its positional accuracy and the reliability of the motion metrics it provides, which are crucial for sports analysis.
    \item We identified specific error patterns in the model that revealed performance biases related to the type of metric, camera view, and subject scale.
\end{itemize}

\section{Related Work}

\textbf{Monocular 2D \& 3D Human Pose Estimation.}
Monocular 2D and 3D human pose estimation is well-developed in computer vision. These techniques are foundational for understanding human movement with a single camera.
2D pose estimation is classified as top-down or bottom-up. Top-down methods \cite{xu2022vitpose, sun2019deep_hrnet} first use a human detector to identify bounding boxes for each person, and then estimate keypoints within each box. This pipeline effectively handles variations in human scale. In contrast, bottom-up approaches \cite{geng2021bottomup, 8765346openpose, cheng2020higherhrnet} detect all keypoints in an image simultaneously and subsequently group them into individual instances, which often results in faster inference speeds. Early methods were predominantly based on convolutional neural networks \cite{sun2019deep_hrnet}. However, recent state-of-the-art models \cite{xu2022vitpose} have increasingly adopted transformer-based architectures, which demonstrate a superior ability to capture an image's global context.
In 3D pose estimation, the primary focus has been on the 2D-to-3D lifting approach, which infers 3D poses from 2D keypoint sequences. Initial advancements in this area demonstrated the effectiveness of temporal convolutions \cite{videopose3d} in modeling motion dynamics. Currently, transformer-based architectures \cite{motionagformer2024, liu2025tcpformer, motionbert2022, strided_transformer, mixste, li2022mhformer, li2023multi_mhformer_journal} that can effectively model spatiotemporal dependencies are the main approach. Recently, diffusion-based methods \cite{Shan_2023_ICCV_D3DP} have also been explored, though they have not surpassed the performance of the leading transformer-based models.

\textbf{Datasets for 3D Human Pose Estimation.}
Advances in 3D pose estimation have relied heavily on large-scale public datasets \cite{ionescu2013h36m, mehta2017mpiinf, 3dpw, Humaneva, TotalCapture, panoptic}. Foundational benchmarks, such as Human3.6M \cite{ionescu2013h36m} and MPI-INF-3DHP \cite{mehta2017mpiinf}, have been crucial for developing these models. However, these datasets primarily feature common daily activities captured in controlled laboratory environments. This results in an absence of characteristic high-speed, dynamic movements.
To address this gap, sports-specific datasets have emerged \cite{AIST++, ASPset, ingwersen2023sportspose, MMSports2024_tanaka_fsjump3d, yeung2025athletepose3d, jiang2024worldpose}. Although these datasets are valuable, they are often limited in their practical utility. They frequently consist of imitated motions \cite{ASPset, ingwersen2023sportspose, yeung2025athletepose3d}, are captured in controlled laboratory environments \cite{AIST++}, and cover only a narrow range of actions \cite{MMSports2024_tanaka_fsjump3d, jiang2024worldpose}. These motions do not fully capture the kinematic complexities of real competitive performances, creating a domain gap that limits model generalization to real-world scenarios. Therefore, there is a critical need for a publicly available dataset featuring authentic, high-speed sports motions captured in a realistic environment to properly train and validate models for sports analysis.

\textbf{Pose Estimation for Motion Analysis.}
The application of pose estimation to motion analysis receives a lot of attention. However, much of the validation work has relied on 2D analysis of simple, controlled movements or multi-view systems to ensure accuracy \cite{hamilton2024comparison_2d_simple, haberkamp2022validity_squat, menychtas2023gait, washabaugh2022comparing_gait, fukushima2024potential, uhlrich2023opencap, torvinen2024evaluation_ski}. Although monocular 3D pose estimation has been applied to complex sports such as soccer \cite{yeung2024autosoccerpose} and golf \cite{ingwersen2023evaluating_golf}, evaluations have primarily focused on positional or angle error \cite{yeung2025athletepose3d}. Therefore, a critical gap remains in validating the practical utility of these models for deriving key performance indicators from real athletic motions.

\section{Method}
\label{sec:method}

In this section, we first explain how to create an AthleticsPose dataset in Sections \ref{sec:data_collection} and \ref{sec:post_processing}. This dataset contains high-speed, ``real'' athletics motions captured on an athletic field. Then, in Section \ref{sec:ps_models}, we describe the monocular pose estimation models evaluated in this study. In Section \ref{sec:ps_validation}, we explain the purpose and method of evaluating these models. Lastly, we address the specific motion analysis to determine the usefulness of monocular pose estimation for sports motion analysis in Section \ref{sec:motion_analysis}.

\subsection{Data Collection}
\label{sec:data_collection}

To create a dataset of real athletics motions for sports-specific pose estimation, we conducted a data collection experiment. A total of 24 university student athletes (15 males, and 9 females, aged 19–26 years) participated in this study. The University’s ethics committee approved the data collection, and written informed consent was obtained from all participants. Each participant performed motions related to their specialty athletics event, and these performances were recorded. To provide a realistic environment for the performance, all data was recorded in an outdoor athletic field.

For motion capture, we used optical markerless motion capture system (Theia3D, Theia). We recorded the trials with eight high-speed cameras (Miqus Video, Qualisys Inc.) with HD resolution. Each camera was positioned to minimize occlusions and operated in a markerless setup. The markerless approach is important because it allows us to capture the motions of athletes without restricting them.
To ensure precise temporal alignment of the captured frames, which is important for accurate 3D motion estimation, we synchronized all the cameras using hardware. We used Qualisys Track Manager (Qualisys) with a wand-type calibration kit to calibrate the cameras and establish their accurate spatial relationships.

\subsection{Post-Processing for Dataset}
\label{sec:post_processing}

To ensure the accuracy, consistency, and usability of the dataset, the raw video data and joint coordinate data captured during the experiments were processed in the following steps.

First, we identified the start and end frames of the video at which the coordinates of all 83 joint positions were detected. Next, we conducted consistency checks of bone lengths to improve the quality of the data. This process involved measuring the lengths of key skeletal segments in each frame and identifying those that significantly deviated from the standard. Such deviations could indicate errors in the markerless calculations. These processes improve the quality of the marker data in the dataset.
Then, only the corresponding video frame was clipped from all the camera videos. This process synchronized the marker data and times of all the videos. Finally, the validated, synchronized, 3D joint positions were projected onto 2D images from each of the eight cameras. This projection was performed using the calibrated intrinsic and extrinsic parameters of each camera.

\begin{table*}[t]
  \caption{Comparison of our AthleticsPose dataset with other major 3D pose datasets. Our ``real'' sports-specific dataset was captured from many athletes performing athletic actions on an outdoor field. Note that since AthletePose3D includes FS-Jump3D, we are showing the results without it.}
  \label{tab:dataset_comparison}
  \scalebox{0.95}{
    \begin{tabular}{lcccccccccc}
        \hline
        Dataset & Sports Reality & Markerless & Sync & Subjects & Poses & Keypoints & Environment & Cameras & Frames & FPS \\
        \hline
        Human3.6M \cite{ionescu2013h36m} & - & $\times$ & hw & 11 & 900K & 26 & lab & 4 & 3.6M & 50 \\
        MPI-INF-3DHP \cite{mehta2017mpiinf} & - & $\checkmark$ & hw & 8 & 93K & 28 & lab \& outdoor & 14 & 1.3M & 25/50 \\
        3DPW \cite{3dpw} & - & $\times$ & sw & 7 & 49K & 24 & lab \& outdoor & 1 & 51K & 30 \\
        HumanEva-I \cite{Humaneva} & - & $\times$ & sw & 6 & 78K & 15 & lab & 7 & 280K & 60 \\
        HumanEva-II \cite{Humaneva} & - & $\times$ & hw & 6 & 3K & 15 & lab & 4 & 10K & 60 \\
        TotalCapture \cite{TotalCapture} & - & $\times$ & hw & 5 & 179K & 25 & lab & 8 & 1.9M & 60 \\
        CMU Panoptic \cite{panoptic} & - & $\checkmark$ & hw & 8 & 1.5M & 18 & lab & 31 & 46.5M & 30 \\
        AIST++ \cite{AIST++} & Realistic & $\checkmark$ & sw & 30 & 1.1M & 17 & lab & 9 & 10.1M & 60 \\
        ASPset-510 \cite{ASPset} & Imitated & $\checkmark$ & sw & 17 & 110K & 17 & outdoor & 3 & 330K & 50 \\
        SportsPose \cite{ingwersen2023sportspose} & Imitated & $\checkmark$ & hw & 24 & 117K & 17 & lab \& outdoor & 7 & 1.5M & 90 \\
        FS-Jump3D \cite{MMSports2024_tanaka_fsjump3d} & Realistic & $\checkmark$ & hw & 4 & 6.5K & 83 & ice rink & 12 & 78K & 60 \\
        AthletePose3D (T\&F) \cite{yeung2025athletepose3d} & Imitated & $\checkmark$ & hw & 4 & 159K & 55/83 & lab & 4/8 & 1.2M & 60/120 \\
        AthleticsPose (Ours) & Realistic & $\checkmark$ & hw & 23 & 62.5K & 83 & athletic field & 8 & 500K & 30/60 \\
        \hline
    \end{tabular}
  }
\end{table*}

\subsection{2D \& 3D Pose Estimation Models}
\label{sec:ps_models}

In this study, rather than conducting an in-depth comparison of monocular 2D \& 3D pose estimation models, we investigate their applicability to motion analysis with representative models.

Monocular 2D pose estimators can be classified into two approaches: top-down and bottom-up. Top-down methods first detect each person with a bounding box, then crop the corresponding image region, and finally estimate keypoints. This pipeline can handle large variations in human scale. In contrast, bottom-up methods infer all people's keypoints at once and subsequently group them into individual instances, offering superior inference speed. Since top-down approaches generally achieve greater accuracy, this study uses ViTPose \cite{xu2022vitpose}, a representative top-down method for experiment.

Monocular 3D pose estimation methods can be grouped into two categories: models that directly infer 3D joint coordinates from raw images or video, and 2D-to-3D lifting models that infer 3D joint positions from sequences of 2D keypoints. Recent research has mainly focused on the 2D-to-3D lifting method, which generally performs better. Consequently, this study uses MotionAGFormer \cite{motionagformer2024}, a representative 2D-to-3D lifting model, for all experiments. The models used in this study have shown high performance not only on general benchmark datasets \cite{ionescu2013h36m} but also for sports datasets in \cite{yeung2025athletepose3d}.

\subsection{3D Pose Estimation Validation}
\label{sec:ps_validation}

Our primary goal is to determine the usefulness of the resulting 3D joint coordinates in motion analysis. Therefore, our experimental evaluation focuses on 3D pose estimation models.

For 2D pose estimation model, we evaluate the change in performance due to fine-tuning. However, the final evaluation focuses on its contribution to the 3D estimation results when the 2D model's output is used as input for the 3D model. These were evaluated separately in a previous study \cite{yeung2025athletepose3d}. In particular, since the 3D pose estimation model was based on the assumption of ground-truth 2D input, the conditions of this study are more practical.

When validating the 3D pose estimation model, we assess not only the accuracy of joint locations but also analyze how factors such as action classes, joint types, and camera views influence performance. Since athletes move around a lot within the camera's view in sports videos, the appearance of motion can vary greatly. Therefore, it is important to know which view is best for analyzing a particular motion. This comprehensive study clarifies realistic deployment scenarios for monocular 3D pose estimation in sports motion analytics. Please refer to Section \ref{sec:evaluation_metrics} for more details about evaluation metrics.

\subsection{Specific Motion Analysis}
\label{sec:motion_analysis}

Previous research on pose estimation techniques has focused on evaluating the accuracy of estimating joint positions, velocities, and angles. However, the usefulness of these techniques for analyzing specific motions has yet to be verified.

In sports, it is important not only to estimate position, velocity, and angle but also to use these results to explain an individual's motion characteristics. Therefore, in addition to pose estimation accuracy, we derive task-specific kinematic metrics from predicted joint positions. We evaluate whether these metrics provide useful information for motion analysis. This study uses two sprint-related indicators as case examples:
\begin{enumerate}
  \item \textbf{Knee angle of the supporting leg.} A larger extension angle indicates that the supporting leg can resist impact more effectively and maintain stability.
  \item \textbf{Maximum knee-drive velocity.} Faster reciprocal knee recovery is strongly correlated with higher sprinting speed.
\end{enumerate}
We validate the practical utility of monocular pose estimation for performance diagnostics by quantifying these variables from estimated 3D joints.

\section{AthleticsPose Dataset}
\label{sec:dataset}

In this section, we first provide an overview of our AthleticsPose dataset. This is the first open dataset of ``real'' athletic motion captured on an athletic field. Next, we compare the basic information of our dataset with that of other popular 3D pose datasets. After that, we present a more detailed analysis of our dataset using the same method as \cite{yeung2025athletepose3d, ingwersen2023sportspose}.

\subsection{Dataset Overview}

We show an overview of AthleticsPose Dataset in Figure \ref{fig:dataset_overview}.
The dataset contains eight unique athletics motions: starting dash (SD), sprinting (Sprint), distance running (Running), race walking (Racewalk), hurdling (Hurdle), shot put (Shotput), discus throw (Discus), and javelin throw. There are 83 joint points for each captured frame of each motion. In total, our dataset comprises approximately 500,000 realistic athletic motion frames.

Table \ref{tab:dataset_comparison} shows the results of comparing the existing major datasets with ours. In addition to the basic settings and statistical comparisons addressed in the previous study \cite{ingwersen2023sportspose, MMSports2024_tanaka_fsjump3d, yeung2025athletepose3d}, we added a section on sports motion reality for the sports-related dataset.
As shown in the table, our AthleticsPose dataset differs from others in several ways. While many datasets are captured in laboratory settings or consist of imitated actions, ours focuses on realistic sports in an outdoor athletic field environment.
To the best of our knowledge, AthleticsPose is the first open 3D human pose dataset featuring authentic athletics movements performed by athletes in an outdoor environment. The dataset provides a large number of keypoints for detailed motion analysis.
General datasets have limitations that make them unsuitable for developing models adapted to the complexities of sports, such as rapid and challenging movements. Therefore, our dataset is an important asset for researchers to validate the accuracy and reliability of pose estimation models for real-world sports applications, promoting advancements in sports science and performance analysis.

\begin{figure}[t]
  \centering
  \includegraphics[width=\linewidth]{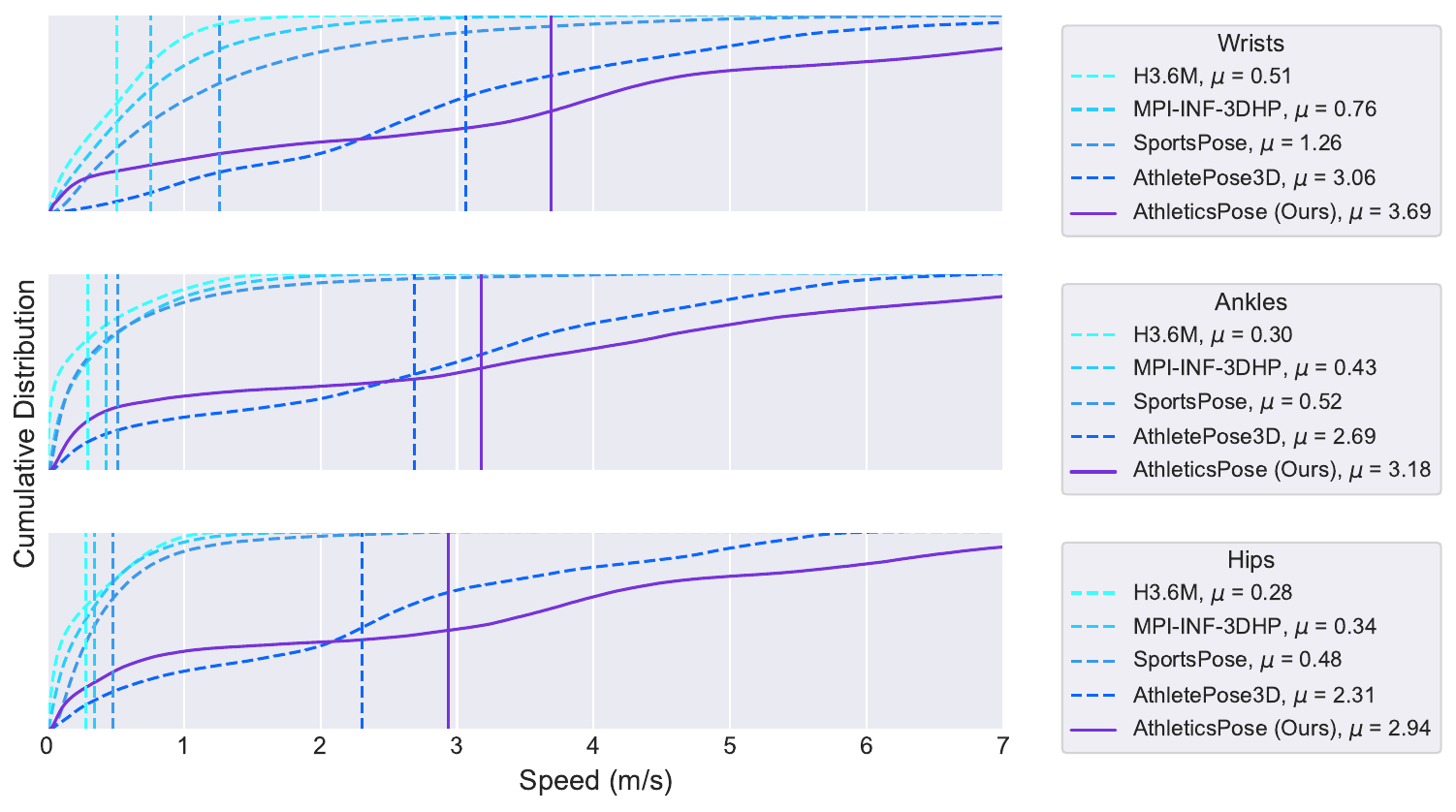}
  \caption{Comparison of Cumulative distribution of joint speeds. The distributions consistently show that our dataset contains motions at significantly higher speeds, as indicated by the right-shifted curves and higher mean values ($\mu$).}
  \label{fig:speed}
\end{figure}

\begin{figure}[t]
  \centering
  \includegraphics[width=\linewidth]{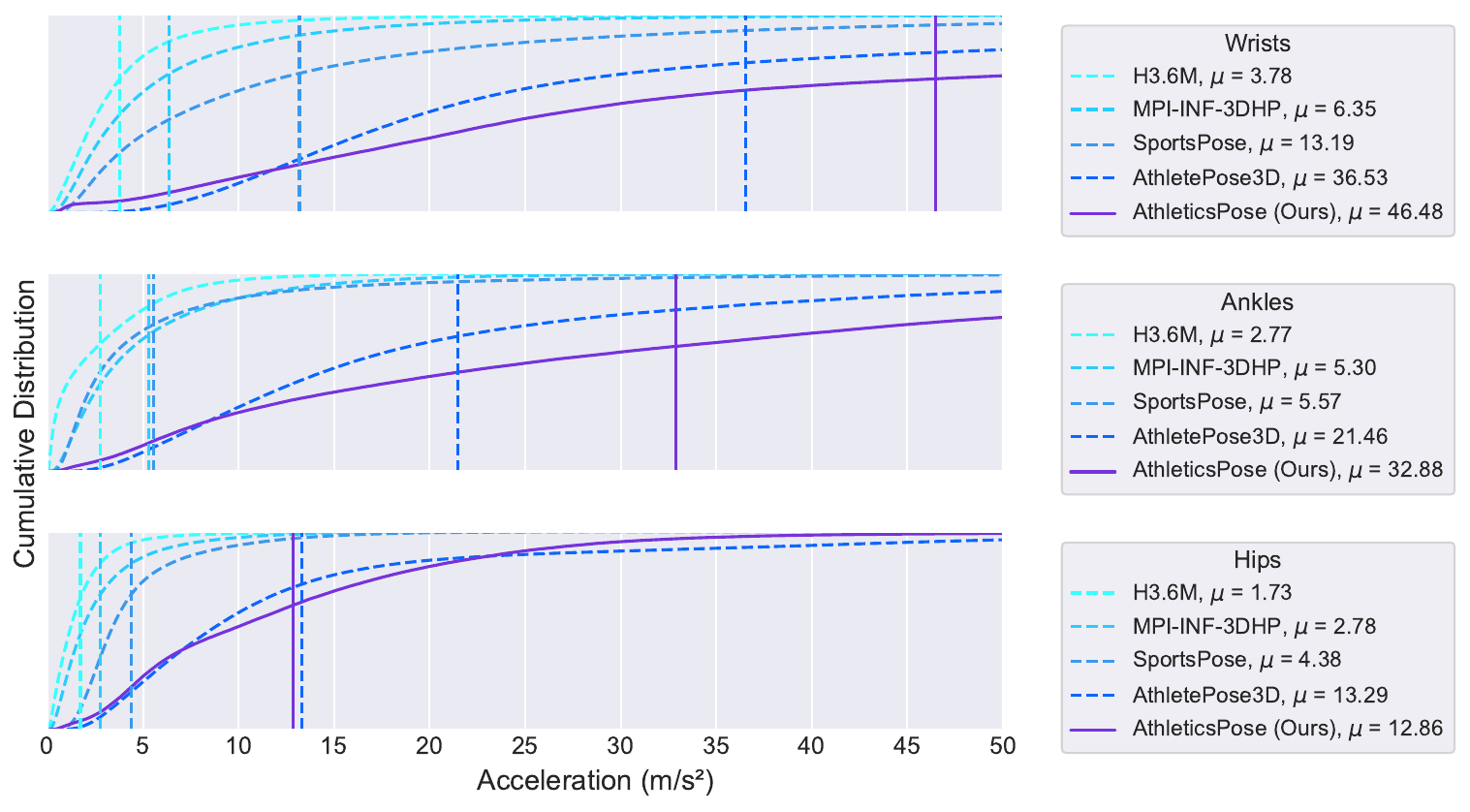}
  \caption{Comparison of Cumulative distribution of joint acceleration. The motions in our dataset are far more dynamic. It contains substantially higher accelerations, which highlights the inclusion of more explosive actions.}
  \label{fig:acceleration}
\end{figure}

\subsection{Detailed Analysis}

Figures \ref{fig:speed} and \ref{fig:acceleration} present the cumulative distribution function (CDF) of speeds and accelerations of key joints (wrists, ankles, and hips) and compare the AthleticsPose dataset to other well-known 3D pose datasets.
The results clearly show that the motions in the AthleticsPose dataset are significantly more dynamic than those in other datasets. The right-shifted curves and higher mean values ($\mu$) in both figures reveal that our dataset consistently covers a wider and higher range of speeds and accelerations.

This is apparent even when compared to the AthletePose3D dataset, which also focuses on sports-related motions. AthleticsPose surpasses AthletePose3D in average speed across all key joints: wrists ($3.69~m/s$ vs. $3.06~m/s$), ankles ($3.18~m/s$ vs. $2.69~m/s$), and hips ($2.94~m/s$ vs. $2.31~m/s$). The difference is even clearer in acceleration: AthleticsPose shows substantially higher acceleration values for the wrists ($46.48~m/s^2$ vs. $36.53~m/s^2$) and ankles ($32.88~m/s^2$ vs. $21.46~m/s^2$), while hip acceleration values are comparable.
Beyond the mean values, the data distribution itself highlights this difference. For instance, approximately $50\%$ of the wrist speeds in the AthletePose3D dataset are at or below $3.0~m/s$. In contrast, this same speed threshold accounts for only about $40\%$ of the AthleticsPose data, indicating a much larger proportion of high-speed movements. These findings highlight the inclusion of more explosive actions in our data.

This significant difference results from the core characteristic of our AthleticsPose dataset: it captures real, high-intensity athletic movements performed by competitive athletes. Unlike datasets comprising daily activities or imitated sports motions, AthleticsPose includes explosive actions such as sprinting, hurdling, and throwing, which inherently involve rapid changes in velocity. These movements represent the upper limit of human physical performance.
Including these high-speed, high-acceleration motions is a crucial advancement in the field. Current 3D pose estimation models often struggle with these extreme dynamics and tend to fail in such conditions. Therefore, AthleticsPose is a valuable and challenging benchmark that allows researchers to evaluate the robustness of their models and develop new methods to accurately capture human motion under challenging athletic conditions.

\section{Experiments}

\subsection{Implementation Details}

We conducted our experiment in Python 3 using Pytorch.
As mentioned in Section \ref{sec:ps_models}, ViTPose \cite{xu2022vitpose} and MotionAGFormer \cite{motionagformer2024} were used for the 2D and 3D pose estimation models, respectively. See Section \ref{sec:model_training} for the detailed training settings of these models.
Throughout all experiments, the dataset was consistently split into training and testing subsets by subject. The test splits were divided based on action, date of capture, and gender to ensure balanced distribution of these factors. The resulting test splits included S00 (Discus and Shotput), S05 (Racewalk), S11, S13, S16 (Running), S12, S21, S22, S23 (SD and Sprint), and S17 and S20 (Hurdle). The remaining subjects are divided into the training split. Please note that we did not include the javelin throw in the evaluation because there was only one subject.
Our dataset, code, and checkpoints are available at \url{https://github.com/SZucchini/AthleticsPose}.


\begin{table*}[t]
  \caption{Quantitative results for 3D pose estimation by actions. All values are MPJPE/P-MPJPE in millimeters. Our model, trained on AthleticsPose, outperforms the baseline models trained on Human3.6M and AthletePose3D significantly.}
  \label{tab:mpjpe_by_action}
  \scalebox{0.81}{
  \begin{tabular}{lcccccccc}
    \toprule
    Model & Avg. & SD & Sprint & Running & Racewalk & Hurdle & Discus & Shotput \\
    \midrule
    Human3.6M (DET pre-trained) & 149.38 / 98.80 & 171.48 / 113.42 & 146.30 / 106.37 & 125.75 / 80.97 & 144.55 / 85.53 & 145.79 / 105.55 & 167.39 / 111.76 & 157.91 / 97.32 \\
    AthletePose3D (GT) & 115.91 / 88.36 & 157.18 / 111.73 & 120.21 / 87.97 & 87.26 / 68.56 & 109.19 / 72.67 & 108.79 / 87.26 & 119.76 / 101.09 & 123.32 / 91.44 \\
    AthleticsPose (DET pre-trained) & 41.99 / 31.40 & 36.03 / 26.92 & 35.39 / 25.57 & 34.62 / 24.11 & 34.64 / 25.29 & 38.16 / 30.17 & 63.57 / 49.00 & 57.76 / 45.05 \\
    AthleticsPose (DET fine-tuned) & 40.13 / 30.33 & 33.43 / 25.12 & 33.85 / 24.15 & 30.61 / 22.50 & 46.35 / 29.53 & 34.43 / 28.14 & 63.18 / 49.32 & 56.73 / 43.66 \\
    AthleticsPose (GT) & 29.45 / 21.44 & 22.23 / 16.40 & 27.12 / 17.72 & 25.37 / 15.71 & 29.41 / 17.29 & 26.46 / 20.02 & 46.63 / 39.23 & 36.76 / 29.10 \\
  \bottomrule
\end{tabular}
}
\end{table*}

\subsection{Pose Estimation Model Training}
\label{sec:model_training}

\subsubsection{\textbf{2D Model Training}}

For the 2D pose estimation model, the base size of ViTPose \cite{xu2022vitpose} was used and trained with MMPose \cite{mmpose2020}, which is a pose estimation framework from OpenMMLab. Since learning from scratch was ineffective, the model was fine-tuned using weights pre-trained by the COCO dataset \cite{cocodataset}. For comparison, we used a model trained on the COCO dataset as the baseline.

During training, 83 marker data points were projected onto the image using the camera's calibration parameters. The largest rectangular area covering these points was then used as the target person's bounding box in the image. Next, the 83 markers were converted to 17 keypoints in the COCO format. Additionally, subjects S03, S07, S14, and S15 were extracted from the training split to create a validation split. Due to the limited size of our dataset, we fine-tuned the model for $60$ epochs with a batch size of $128$. We enabled automatic scaling of the learning rate to match the original settings.

\subsubsection{\textbf{3D Model Training}}

We used the base size of MotionAGFormer \cite{motionagformer2024} for the 3D pose estimation model and set the input frame length to $81$. We trained the model from scratch using our dataset. We used the models trained on the Human3.6M and AthletePose3D as a baseline for comparison. 

For training, we designed two conditions: one with 2D ground-truth input (GT) and one with input from the 2D pose estimation model (DET). In the first condition, the 83 markers in the dataset were projected into 2D space and converted directly into 17 keypoints in the Human3.6M format. In the latter case, the 17 keypoints in COCO format output by the 2D pose estimation model were converted to Human3.6M format and used as input. Since the AthletePose3D model was the only publicly available model for the GT condition, it was evaluated in that condition.
In the real world, the DET condition is more practical because obtaining accurate 2D ground-truth is difficult. Our study differs from previous work \cite{yeung2025athletepose3d} in that we evaluate the performance of 3D pose estimation under the DET condition, which is closer to realistic sports applications.
For the DET condition, we used two 2D pose estimation models. One used pre-trained ViTPose from the COCO dataset as input (DET pre-trained), and the other used fine-tuned ViTPose from our dataset as input (DET fine-tuned). In summary, the training conditions for 3D pose estimation can be divided into three categories: GT, DET pre-trained, and DET fine-tuned.

Following previous studies \cite{ci2019optimizing_lcnpose, videopose3d, motionagformer2024}, the 3D label data was converted from the camera coordinate system to the pixel coordinate system. In addition, the 2D input and 3D label data were both converted to relative coordinates with respect to the root joint (the pelvis in the Human3.6M format). Furthermore, all data was normalized to the range of $[-1, 1]$. Please note that while some studies used image resolution-based normalization, this method changes the shape of the input pose depending on the aspect ratio and image size. Therefore, absolute value-based normalization of pixel coordinates was used in this experiment.
Except for increasing the number of epochs from $60$ to $100$ to allow for convergence of learning, all other conditions were inherited from the default settings of MotionAGFormer.

\subsection{Evaluation Metrics}
\label{sec:evaluation_metrics}

\subsubsection{\textbf{For 2D Pose Estimation}}
To evaluate the performance of 2D pose estimation, we used the standard metric of mean Average Precision (mAP), which is widely used in benchmarks such as the COCO keypoint detection task. The evaluation is based on Object Keypoint Similarity (OKS), which measures the normalized distance between a predicted keypoint and its corresponding ground-truth. OKS is defined as follows:
\begin{equation}
    \text{OKS} = \frac{\sum_{i} \exp(-d_i^2 / 2s^2k_i^2) \cdot \delta(v_i > 0)}{\sum_{i} \delta(v_i > 0)}
\end{equation}
where $d_i$ is the Euclidean distance between the predicted and ground-truth keypoints, $v_i$ is the visibility flag of the ground-truth keypoint, $s$ is the object scale, and $k_i$ is a per-keypoint constant that controls falloff. A prediction is considered a true positive if its OKS with the ground-truth is above a certain threshold.

Average Precision (AP) is calculated by averaging the precision values across multiple recall levels. The final mAP score is computed by averaging the AP over a range of OKS thresholds from 0.50 to 0.95 with a step size of 0.05. In our evaluation, we report the primary mAP score as the key metric for model performance.

\subsubsection{\textbf{For 3D Pose Estimation}}
We evaluate 3D pose estimation using two widely employed metrics that are considered fundamental in the field \cite{motionagformer2024, motionbert2022, videopose3d}.
The primary metric is the Mean Per Joint Position Error (MPJPE). It provides a simple, clear measure of the average Euclidean distance between predicted 3D joint coordinates and their corresponding ground-truth locations. This error is averaged over all joints in a given pose. Due to its direct measurement of absolute localization accuracy, MPJPE is a standard benchmark in 3D pose estimation studies.

We also report the Procrustes-aligned MPJPE (P-MPJPE). This metric aligns the predicted 3D pose with the ground-truth pose via a similarity transformation, or Procrustes analysis. This process removes differences in global translation, rotation, and scaling. After alignment, the MPJPE is calculated. This protocol is a standard evaluation procedure that provides an accurate evaluation of the predicted pose structure, independent of its global position or orientation. It provides a better comparison of a model's ability to correctly estimate the body's configuration.

\subsubsection{\textbf{For Specific Motion Analysis}}
To validate the effectiveness of 3D pose estimation for specific motion analysis, we calculate and evaluate the knee angles of the supporting leg and the maximum knee-drive velocity, with a focus on sprinting, as discussed in Section \ref{sec:motion_analysis}.

For the analysis of the knee angles of the supporting leg, we calculated the knee angle for each leg using the predicted 3D joint coordinates of the hip, knee, and ankle. Then, we smoothed the calculated knee angles using a low-pass filter. To identify the supporting leg and timing, the moment of minimum knee distance is detected for each step. The leg with the larger knee angle (i.e., the more extended leg) in that frame is identified as the supporting leg, and its angle is extracted for analysis.
To evaluate maximum knee-drive velocity, the distance between the knees is calculated using the predicted joint coordinates and smoothed with a low-pass filter. Next, the velocity is calculated from the distance. Finally, the maximum velocity during knee drive is obtained by detecting the peak velocity value. Each peak value indicates the maximum velocity of a single step. To evaluate their errors, the angles and velocities are compared with the results calculated using ground-truth joint positions.

\subsection{Results}

In this section, we first present the results of fine-tuning the 2D pose estimation model. Then, we describe the basic evaluation results of 3D pose estimation models trained on our AthleticsPose dataset. Following that, we show a more detailed evaluation of 3D pose estimation regarding camera view and input target scale, which is important for sports motion analysis. Finally, we demonstrate the results of a specific motion analysis of sprinting using the 3D pose estimation model.

For fine-tuning of the 2D pose estimation model, the mAP increased from $0.875$ to $0.979$ compared to the baseline model. Although the baseline model had high performance, fine-tuning increased its performance almost to the upper limit. This result suggests that fine-tuning is an effective approach for 2D pose estimation when the dataset for a specific domain is limited. However, our main interest is not improving the performance of 2D pose estimation but clarifying how it contributes to 3D pose estimation.

For a basic evaluation of 3D pose estimation, Table \ref{tab:mpjpe_by_action} shows the evaluation results for each action based on the type of training data. First, in terms of dataset type, the model trained on the Human3.6M dataset was compared with the model trained on the AthleticsPose dataset. The MPJPE decreased from $149.38~mm$ to $41.99~mm$, while the P-MPJPE decreased from $98.90~mm$ to $31.40~mm$. Thus, using the sports-specific dataset improved performance by approximately $70\%$ on average. 

We then compared the two sports-specific datasets, AthletePose3D and our AthleticsPose, under the GT input condition. Training on our dataset achieved a significantly lower MPJPE of $29.45~mm$ compared to AthletePose3D's MPJPE of $115.91~mm$. This improvement is due to the inclusion of ``Realistic'' motions in our dataset as opposed to the ``Imitated'' motions in AthletePose3D. This limitation is present even in actions that are included in both datasets, such as running, discus, and shotput. A model trained on imitated motions fails to generalize effectively to authentic motions in our test set. This emphasizes the importance of capturing real motions in sports.

Next, we compared the different types of 2D input (GT or DET). The GT condition showed the best performance. Under GT conditions, 2D input is unaffected by motion blur or occlusion, which results in higher performance. Comparing models that estimate 2D input (DET pre-trained or fine-tuned) revealed that the DET fine-tuned slightly outperformed the DET pre-trained. These results suggest that the pre-trained DET performs well enough and that the 3D pose estimation model can reduce the effects of errors in 2D estimation. These results also indicate that the difference between GT and DET is significant and cannot be resolved by fine-tuning alone. The results by action show that actions including periodic motions, such as running, had smaller errors than actions not including periodic motions, such as throwing. This suggests that the model can recognize common features in similar periodic actions.

To evaluate the performance by action in more detail, Table \ref{tab:mpjpe_by_joint} shows the results of evaluating MPJPE by major joint for each action. Please note that all results presented below are evaluation results using AthleticsPose (DET fine-tuned), unless otherwise mentioned, because this is a practical setting.
In general, the farther a joint is from the root joint (the pelvis, which is near the hip), the larger the error. This is because monocular 3D pose estimation estimates the relative coordinates of joints in relation to the root joint. A more notable result is the increase in upper limb (Elbow and Wrist) errors in throwing actions (Discus and Shotput). Accurate analysis of the motion of the upper limbs during these actions is important, but it can be expected that it would be difficult with such a large margin error. Therefore, the evaluation of 3D pose estimation for motion analysis in this study deals with motions other than throwing motions.

Then, Table \ref{tab:mpjpe_by_cam} shows the evaluation results by camera view to explore conditions that are more suitable for sports motion analysis. Here, we focused on actions other than throwing because the position on the screen is more variable for these actions. For this experiment, camera views were classified into two categories: side and front/back. This classification was based on scale changes of the target in the image. See Figure \ref{fig:dataset_overview} for examples of each view.
As shown in the table, the side view has lower error ($31.42~mm$ MPJPE) than the front/back view ($36.34~mm$ MPJPE). A similar trend is observed for P-MPJPE. These results suggest that the model's performance is sensitive to the scale of the subject in the image. The side view maintains a relatively constant subject scale, whereas the front/back view involves significant scale variations when athletes move toward or away from the camera. These findings underscore the importance of selecting an appropriate camera view to ensure reliable sports motion analysis, especially for running-related motions in this case.

\begin{table}[t]
  \caption{Per-joint analysis for each action (MPJPE in millimeters). The results show that errors are particularly large in upper-body joints during throwing motions.}
  \label{tab:mpjpe_by_joint}
  \scalebox{0.95}{
  \begin{tabular}{lcccccc}
    \toprule
    Action & Hip & Knee & Ankle & Shoulder & Elbow & Wrist \\
    \midrule
    Avg. & 12.88 & 40.26 & 50.59 & 38.10 & 53.81 & 70.17 \\
    SD & 8.00 & 37.22 & 42.69 & 33.10 & 41.02 & 49.65 \\
    Sprint & 12.14 & 33.23 & 48.12 & 35.53 & 39.65 & 47.63 \\
    Running & 9.79 & 35.07 & 41.16 & 29.59 & 40.32 & 47.36 \\
    Racewalk & 17.15 & 40.55 & 58.38 & 50.58 & 68.33 & 65.06 \\
    Hurdle & 11.32 & 34.76 & 45.09 & 30.15 & 46.56 & 70.27 \\
    Discus & 18.27 & 52.52 & 73.63 & 57.25 & 91.10 & 134.97 \\
    Shotput & 16.94 & 55.65 & 66.44 & 51.68 & 75.80 & 100.49 \\
  \bottomrule
\end{tabular}
}
\end{table}

To further investigate the model's dependency on target scale, we analyzed its performance by categorizing the input data according to bounding box height. Specifically, we calculated a relative scale for each frame by normalizing the bounding box height. We treated the maximum height observed in all input frames as $1.0$. We defined ``Large'' as inputs with a relative scale of $0.67$ or greater, ``Medium'' as inputs with a relative scale above $0.33$, and ``Small'' as inputs with a relative scale of $0.33$ or less.
Table \ref{tab:mpjpe_by_scale} clearly shows a strong correlation between input scale and estimation accuracy. Performance degrades as the target's scale decreases. For example, the MPJPE increases from $33.27~mm$ for ``Large'' inputs to $42.63~mm$ for ``Small'' inputs. A similar trend is observed for the P-MPJPE.
This finding provides a clearer explanation for the results observed in our camera view analysis. The higher error in the front and back views is mainly caused by smaller targets moving within the scene. Although the training data included examples from all scale categories, the model has difficulty maintaining high accuracy for smaller targets. One likely reason is the degraded accuracy of the initial 2D keypoint detection for these smaller, lower-resolution targets. Additionally, the 3D model itself may have difficulty accurately reasoning about pose structure from such small-scale inputs. For sports motion analysis applications, where precision is essential, these results underscore the importance of camera placement strategies that capture targets at an appropriate scale.

\begin{table}
  \caption{3D pose estimation error (MPJPE and P-MPJPE in millimeters) analyzed by camera view.}
  \label{tab:mpjpe_by_cam}
  \begin{tabular}{lcc}
    \toprule
    Camera view & MPJPE & P-MPJPE \\
    \midrule
    Side view & 31.42 & 24.12 \\
    Front / Back view & 36.34 & 26.53 \\
  \bottomrule
\end{tabular}
\vspace{10pt}
\end{table}

\begin{table}
  \caption{3D pose estimation error (MPJPE and P-MPJPE in millimeters) analyzed by input scale.}
  \label{tab:mpjpe_by_scale}
  \begin{tabular}{lcc}
    \toprule
    Input scale & MPJPE & P-MPJPE \\
    \midrule
    Large & 33.27 & 24.74 \\
    Medium & 37.15 & 27.57 \\
    Small & 42.63 & 32.58 \\
  \bottomrule
\end{tabular}
\end{table}

\begin{figure}[t]
  \centering
  \includegraphics[width=0.9\linewidth]{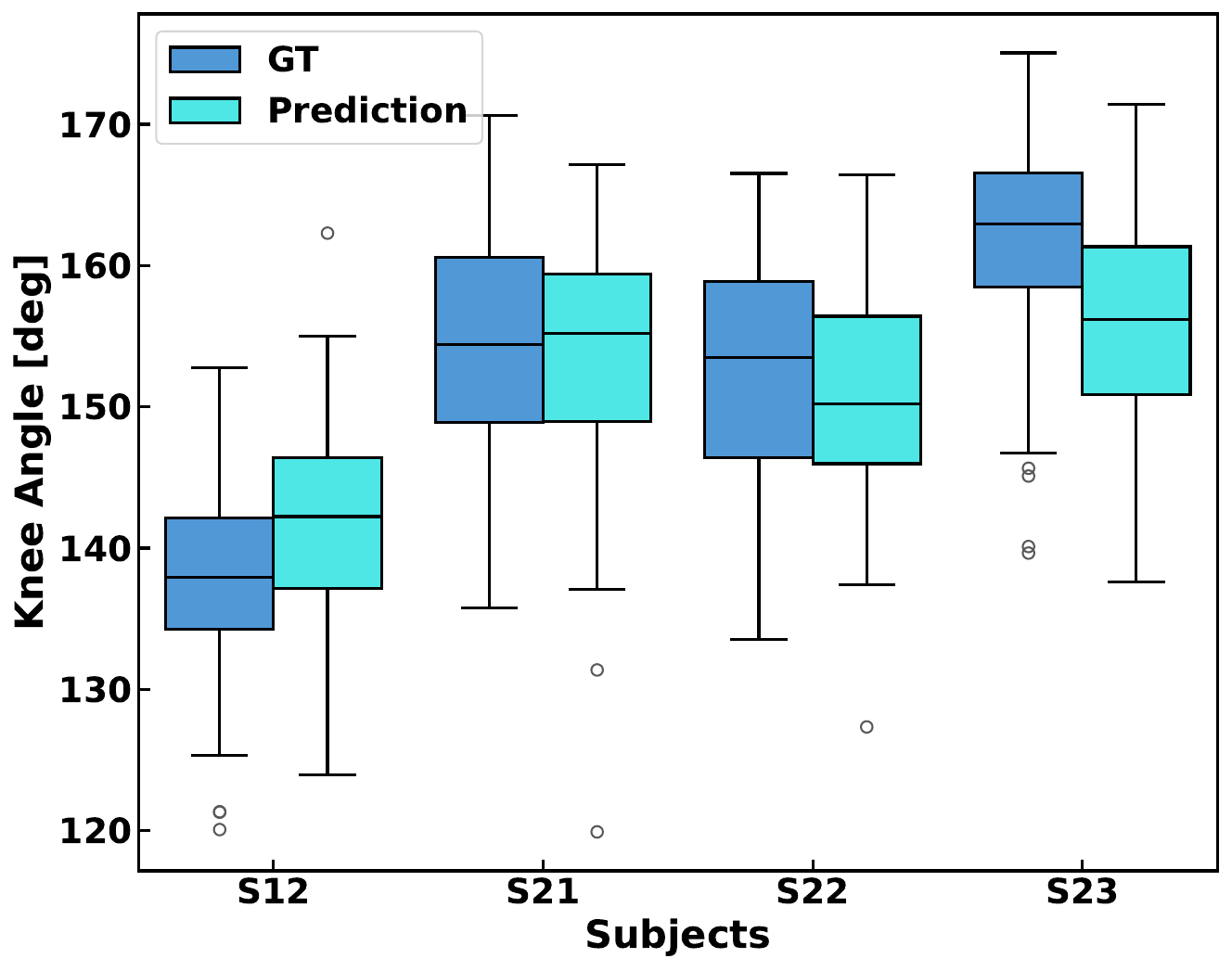}
  \caption{Comparison of GT and predicted knee angles by subject, illustrating the model's ability to capture individual differences.}
  \label{fig:angle_boxplot}
\end{figure}

Finally, we present the results of the specific motion analysis. First, the RMSE of the knee angle calculated from the ground-truth and the knee angle calculated from the predicted joint position was $6.43^\circ$ for the supporting leg. When calculated by camera view, the RMSE was $4.82^\circ$ for the side view and $7.18^\circ$ for the front/back view. These results suggest that the knee angle can be accurately evaluated and varies depending on the camera view.
An analysis of variance (ANOVA) and subsequent Tukey's HSD post-hoc test confirmed that 3D pose estimation successfully captures subject-specific patterns. Significant differences were found in both ground-truth (GT) and predicted knee angles among subjects ($p < 0.05$).
However, the analysis revealed two biases, as can be seen in the box plot in Figure \ref{fig:angle_boxplot}. First, the predicted angles underestimated the magnitude of inter-subject differences (e.g., the mean difference between S12 and S23 was $14.66^\circ$ predicted vs. $24.07^\circ$ GT). Second, the model created a statistical distinction where none existed in the GT. Subjects S21 and S22 were not significantly different ($p = 0.089$), yet the predicted angles showed a significant difference ($p = 0.003$).
In summary, these findings demonstrate that while 3D pose estimation is a valuable tool for subject-specific motion analysis, its practical application requires careful consideration of quantitative biases.

\begin{figure}[t]
  \centering
  \includegraphics[width=0.9\linewidth]{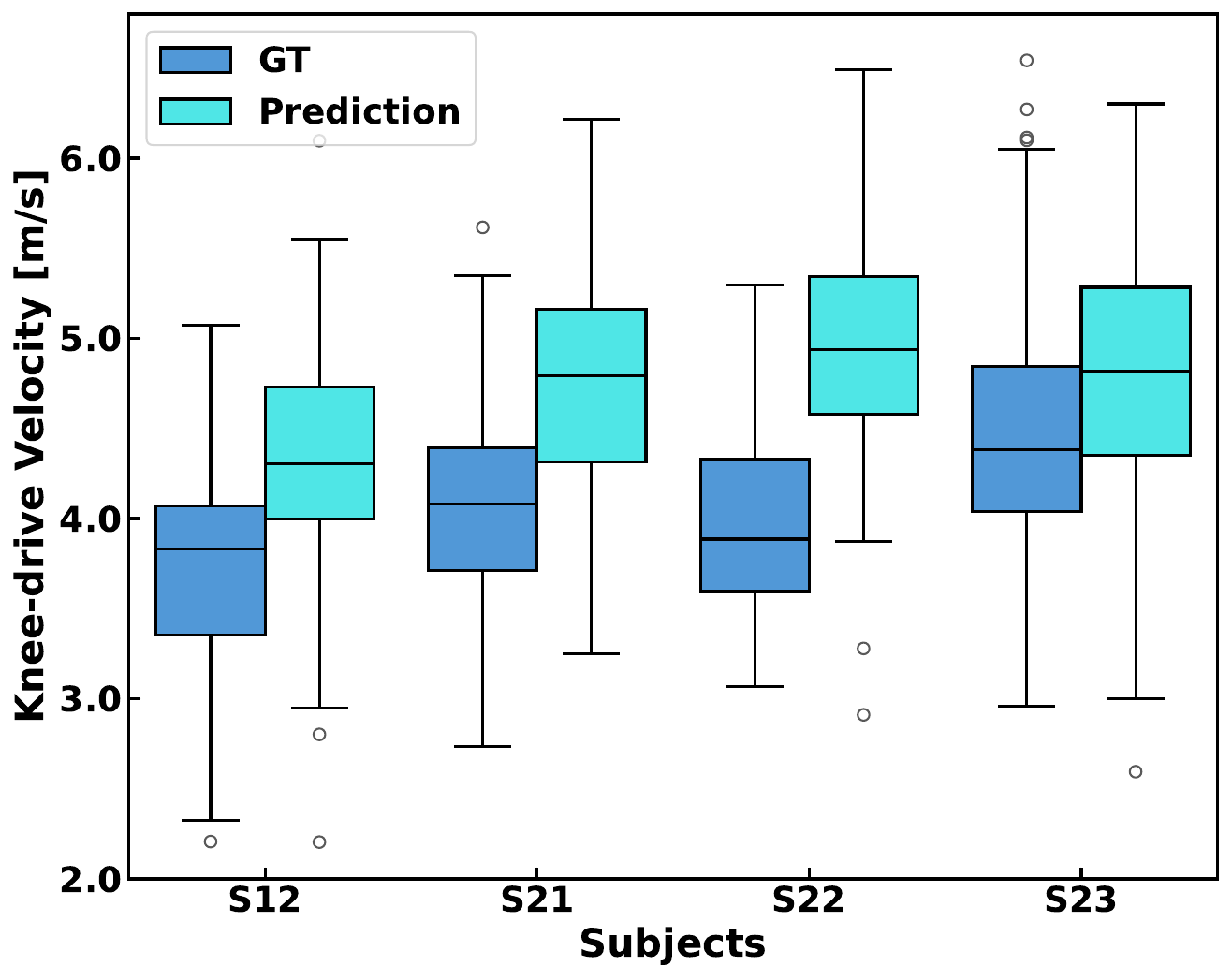}
  \caption{Comparison of GT and predicted knee-drive velocity by subject.}
  \label{fig:velocity_boxplot}
\end{figure}

The overall RMSE for the knee-drive velocity metric was $0.77~m/s$. Unlike the knee angle results, the error was slightly higher for the side view ($0.79~m/s$) than for the front/back view ($0.76~m/s$).
As shown in the box plot in Figure \ref{fig:velocity_boxplot}, the predicted results exhibited consistent overestimation across all subjects. Although an ANOVA revealed statistically significant differences between subjects in both the GT and predicted values (p < 0.05), a subsequent Tukey's HSD post-hoc test showed that the model failed to replicate these differences accurately.
This suggests that the model is limited in its utility for detailed, subject-specific analyses of this velocity-based metric because it does not accurately preserve true relational differences between individuals.

\section{Conclusion}
In this paper, we introduced the AthleticsPose dataset, which contains authentic athletics motions captured on an athletic field. This dataset is useful for advancing the development and validation of sports-specific 3D pose estimation models.
Using the dataset, we revealed that the critical importance of authentic sports motion data, as we discovered that the model trained on imitated sports motions cannot effectively generalize to real-world sports motions.
In addition, we demonstrated that the practical application of 3D pose estimation is highly sensitive to environmental factors. Performance varies significantly with camera view and input scale, highlighting the need for careful camera placement in practice and for the development of more scale-robust models in future work.

Furthermore, our analysis of specific kinematic metrics for sprinting produced variable results. While estimated knee angles of the supporting leg showed potential for capturing subject-specific differences, analyzing knee-drive velocity proved more challenging. The model failed to replicate the true pattern of inter-subject differences for this velocity metric. This suggests that the current level of joint position error (MPJPE) limits the model's utility for analyzing such high-speed movements. Overall, our work clarifies the promising potential and current limitations of using monocular 3D pose estimation for detailed sports motion analysis.

\begin{acks}
This work was supported by the Public Interest Incorporated Foundation, the Yamaha Motor Foundation for Sports, and JST BOOST Japan, Grant Number JPMJBS2422.
Additionally, we received kind support from the Nagoya University Track \& Field Club during the data collection process. We would like to express our sincere gratitude to all the athletes on the team.
\end{acks}
\bibliographystyle{ACM-Reference-Format}
\bibliography{sample-base}


\begin{thebibliography}{43}


\ifx \showCODEN    \undefined \def \showCODEN     #1{\unskip}     \fi
\ifx \showISBNx    \undefined \def \showISBNx     #1{\unskip}     \fi
\ifx \showISBNxiii \undefined \def \showISBNxiii  #1{\unskip}     \fi
\ifx \showISSN     \undefined \def \showISSN      #1{\unskip}     \fi
\ifx \showLCCN     \undefined \def \showLCCN      #1{\unskip}     \fi
\ifx \shownote     \undefined \def \shownote      #1{#1}          \fi
\ifx \showarticletitle \undefined \def \showarticletitle #1{#1}   \fi
\ifx \showURL      \undefined \def \showURL       {\relax}        \fi
\providecommand\bibfield[2]{#2}
\providecommand\bibinfo[2]{#2}
\providecommand\natexlab[1]{#1}
\providecommand\showeprint[2][]{arXiv:#2}

\bibitem[{Cao} et~al\mbox{.}(2019)]%
        {8765346openpose}
\bibfield{author}{\bibinfo{person}{Z. {Cao}}, \bibinfo{person}{G. {Hidalgo Martinez}}, \bibinfo{person}{T. {Simon}}, \bibinfo{person}{S. {Wei}}, {and} \bibinfo{person}{Y.~A. {Sheikh}}.} \bibinfo{year}{2019}\natexlab{}.
\newblock \showarticletitle{OpenPose: Realtime Multi-Person 2D Pose Estimation using Part Affinity Fields}.
\newblock \bibinfo{journal}{\emph{IEEE Transactions on Pattern Analysis and Machine Intelligence}} (\bibinfo{year}{2019}).
\newblock


\bibitem[Cheng et~al\mbox{.}(2020)]%
        {cheng2020higherhrnet}
\bibfield{author}{\bibinfo{person}{Bowen Cheng}, \bibinfo{person}{Bin Xiao}, \bibinfo{person}{Jingdong Wang}, \bibinfo{person}{Honghui Shi}, \bibinfo{person}{Thomas~S Huang}, {and} \bibinfo{person}{Lei Zhang}.} \bibinfo{year}{2020}\natexlab{}.
\newblock \showarticletitle{Higherhrnet: Scale-aware representation learning for bottom-up human pose estimation}. In \bibinfo{booktitle}{\emph{Proceedings of the IEEE/CVF conference on computer vision and pattern recognition}}. \bibinfo{pages}{5386--5395}.
\newblock


\bibitem[Ci et~al\mbox{.}(2019)]%
        {ci2019optimizing_lcnpose}
\bibfield{author}{\bibinfo{person}{Hai Ci}, \bibinfo{person}{Chunyu Wang}, \bibinfo{person}{Xiaoxuan Ma}, {and} \bibinfo{person}{Yizhou Wang}.} \bibinfo{year}{2019}\natexlab{}.
\newblock \showarticletitle{Optimizing Network Structure for 3D Human Pose Estimation}. In \bibinfo{booktitle}{\emph{Proceedings of the IEEE International Conference on Computer Vision}}. \bibinfo{pages}{2262--2271}.
\newblock


\bibitem[Contributors(2020)]%
        {mmpose2020}
\bibfield{author}{\bibinfo{person}{MMPose Contributors}.} \bibinfo{year}{2020}\natexlab{}.
\newblock \bibinfo{title}{OpenMMLab Pose Estimation Toolbox and Benchmark}.
\newblock \bibinfo{howpublished}{\url{https://github.com/open-mmlab/mmpose}}.
\newblock


\bibitem[Fukushima et~al\mbox{.}(2024)]%
        {fukushima2024potential}
\bibfield{author}{\bibinfo{person}{Takashi Fukushima}, \bibinfo{person}{Patrick Blauberger}, \bibinfo{person}{Tiago Guedes~Russomanno}, {and} \bibinfo{person}{Martin Lames}.} \bibinfo{year}{2024}\natexlab{}.
\newblock \showarticletitle{The potential of human pose estimation for motion capture in sports: a validation study}.
\newblock \bibinfo{journal}{\emph{Sports Engineering}} \bibinfo{volume}{27}, \bibinfo{number}{1} (\bibinfo{year}{2024}), \bibinfo{pages}{19}.
\newblock


\bibitem[Geng et~al\mbox{.}(2021)]%
        {geng2021bottomup}
\bibfield{author}{\bibinfo{person}{Zigang Geng}, \bibinfo{person}{Ke Sun}, \bibinfo{person}{Bin Xiao}, \bibinfo{person}{Zhaoxiang Zhang}, {and} \bibinfo{person}{Jingdong Wang}.} \bibinfo{year}{2021}\natexlab{}.
\newblock \showarticletitle{Bottom-up human pose estimation via disentangled keypoint regression}. In \bibinfo{booktitle}{\emph{Proceedings of the IEEE/CVF conference on computer vision and pattern recognition}}. \bibinfo{pages}{14676--14686}.
\newblock


\bibitem[Haberkamp et~al\mbox{.}(2022)]%
        {haberkamp2022validity_squat}
\bibfield{author}{\bibinfo{person}{Lucas~D Haberkamp}, \bibinfo{person}{Micah~C Garcia}, {and} \bibinfo{person}{David~M Bazett-Jones}.} \bibinfo{year}{2022}\natexlab{}.
\newblock \showarticletitle{Validity of an artificial intelligence, human pose estimation model for measuring single-leg squat kinematics}.
\newblock \bibinfo{journal}{\emph{Journal of Biomechanics}}  \bibinfo{volume}{144} (\bibinfo{year}{2022}), \bibinfo{pages}{111333}.
\newblock


\bibitem[Hamilton et~al\mbox{.}(2024)]%
        {hamilton2024comparison_2d_simple}
\bibfield{author}{\bibinfo{person}{Rebecca~I Hamilton}, \bibinfo{person}{Zornitza Glavcheva-Laleva}, \bibinfo{person}{Md~Imdadul~Haque Milon}, \bibinfo{person}{Yeshwin Anil}, \bibinfo{person}{Jenny Williams}, \bibinfo{person}{Peter Bishop}, {and} \bibinfo{person}{Catherine Holt}.} \bibinfo{year}{2024}\natexlab{}.
\newblock \showarticletitle{Comparison of computational pose estimation models for joint angles with 3D motion capture}.
\newblock \bibinfo{journal}{\emph{Journal of Bodywork and Movement Therapies}}  \bibinfo{volume}{40} (\bibinfo{year}{2024}), \bibinfo{pages}{315--319}.
\newblock


\bibitem[Ingwersen et~al\mbox{.}(2023a)]%
        {ingwersen2023evaluating_golf}
\bibfield{author}{\bibinfo{person}{Christian~Keilstrup Ingwersen}, \bibinfo{person}{Janus~N{\o}rtoft Jensen}, \bibinfo{person}{Morten~Rieger Hannemose}, {and} \bibinfo{person}{Anders~Bjorholm Dahl}.} \bibinfo{year}{2023}\natexlab{a}.
\newblock \showarticletitle{Evaluating current state of monocular 3D pose models for golf}. In \bibinfo{booktitle}{\emph{Northern Lights Deep Learning Workshop 2023}}. Septentrio Academic Publishing.
\newblock


\bibitem[Ingwersen et~al\mbox{.}(2023b)]%
        {ingwersen2023sportspose}
\bibfield{author}{\bibinfo{person}{Christian~Keilstrup Ingwersen}, \bibinfo{person}{Christian~M{\o}ller Mikkelstrup}, \bibinfo{person}{Janus~N{\o}rtoft Jensen}, \bibinfo{person}{Morten~Rieger Hannemose}, {and} \bibinfo{person}{Anders~Bjorholm Dahl}.} \bibinfo{year}{2023}\natexlab{b}.
\newblock \showarticletitle{Sportspose-a dynamic 3d sports pose dataset}. In \bibinfo{booktitle}{\emph{Proceedings of the IEEE/CVF Conference on Computer Vision and Pattern Recognition}}. \bibinfo{pages}{5219--5228}.
\newblock


\bibitem[Ionescu et~al\mbox{.}(2013)]%
        {ionescu2013h36m}
\bibfield{author}{\bibinfo{person}{Catalin Ionescu}, \bibinfo{person}{Dragos Papava}, \bibinfo{person}{Vlad Olaru}, {and} \bibinfo{person}{Cristian Sminchisescu}.} \bibinfo{year}{2013}\natexlab{}.
\newblock \showarticletitle{Human3. 6m: Large scale datasets and predictive methods for 3d human sensing in natural environments}.
\newblock \bibinfo{journal}{\emph{IEEE transactions on pattern analysis and machine intelligence}} \bibinfo{volume}{36}, \bibinfo{number}{7} (\bibinfo{year}{2013}), \bibinfo{pages}{1325--1339}.
\newblock


\bibitem[Jiang et~al\mbox{.}(2024)]%
        {jiang2024worldpose}
\bibfield{author}{\bibinfo{person}{Tianjian Jiang}, \bibinfo{person}{Johsan Billingham}, \bibinfo{person}{Sebastian M{\"u}ksch}, \bibinfo{person}{Juan Zarate}, \bibinfo{person}{Nicolas Evans}, \bibinfo{person}{Martin~R Oswald}, \bibinfo{person}{Marc Polleyfeys}, \bibinfo{person}{Otmar Hilliges}, \bibinfo{person}{Manuel Kaufmann}, {and} \bibinfo{person}{Jie Song}.} \bibinfo{year}{2024}\natexlab{}.
\newblock \showarticletitle{WorldPose: A world cup dataset for global 3D human pose estimation}. In \bibinfo{booktitle}{\emph{European Conference on Computer Vision}}. Springer, \bibinfo{pages}{343--362}.
\newblock


\bibitem[Joo et~al\mbox{.}(2015)]%
        {panoptic}
\bibfield{author}{\bibinfo{person}{Hanbyul Joo}, \bibinfo{person}{Hao Liu}, \bibinfo{person}{Lei Tan}, \bibinfo{person}{Lin Gui}, \bibinfo{person}{Bart Nabbe}, \bibinfo{person}{Iain Matthews}, \bibinfo{person}{Takeo Kanade}, \bibinfo{person}{Shohei Nobuhara}, {and} \bibinfo{person}{Yaser Sheikh}.} \bibinfo{year}{2015}\natexlab{}.
\newblock \showarticletitle{Panoptic studio: A massively multiview system for social motion capture}. In \bibinfo{booktitle}{\emph{Proceedings of the IEEE International Conference on Computer Vision}}. \bibinfo{pages}{3334--3342}.
\newblock


\bibitem[Li et~al\mbox{.}(2021)]%
        {AIST++}
\bibfield{author}{\bibinfo{person}{Ruilong Li}, \bibinfo{person}{Sha Yang}, \bibinfo{person}{David~A. Ross}, {and} \bibinfo{person}{Angjoo Kanazawa}.} \bibinfo{year}{2021}\natexlab{}.
\newblock \showarticletitle{AI Choreographer: Music Conditioned 3D Dance Generation with AIST++}.
\newblock \bibinfo{journal}{\emph{2021 IEEE/CVF International Conference on Computer Vision (ICCV)}} (\bibinfo{year}{2021}), \bibinfo{pages}{13381--13392}.
\newblock
\urldef\tempurl%
\url{https://api.semanticscholar.org/CorpusID:236882798}
\showURL{%
\tempurl}


\bibitem[Li et~al\mbox{.}(2022a)]%
        {strided_transformer}
\bibfield{author}{\bibinfo{person}{Wenhao Li}, \bibinfo{person}{Hong Liu}, \bibinfo{person}{Runwei Ding}, \bibinfo{person}{Mengyuan Liu}, \bibinfo{person}{Pichao Wang}, {and} \bibinfo{person}{Wenming Yang}.} \bibinfo{year}{2022}\natexlab{a}.
\newblock \showarticletitle{Exploiting temporal contexts with strided transformer for 3d human pose estimation}.
\newblock \bibinfo{journal}{\emph{IEEE Transactions on Multimedia}}  \bibinfo{volume}{25} (\bibinfo{year}{2022}), \bibinfo{pages}{1282--1293}.
\newblock


\bibitem[Li et~al\mbox{.}(2023)]%
        {li2023multi_mhformer_journal}
\bibfield{author}{\bibinfo{person}{Wenhao Li}, \bibinfo{person}{Hong Liu}, \bibinfo{person}{Hao Tang}, {and} \bibinfo{person}{Pichao Wang}.} \bibinfo{year}{2023}\natexlab{}.
\newblock \showarticletitle{Multi-Hypothesis Representation Learning for Transformer-Based 3D Human Pose Estimation}.
\newblock \bibinfo{journal}{\emph{Pattern Recognition}}  \bibinfo{volume}{141} (\bibinfo{year}{2023}), \bibinfo{pages}{109631}.
\newblock


\bibitem[Li et~al\mbox{.}(2022b)]%
        {li2022mhformer}
\bibfield{author}{\bibinfo{person}{Wenhao Li}, \bibinfo{person}{Hong Liu}, \bibinfo{person}{Hao Tang}, \bibinfo{person}{Pichao Wang}, {and} \bibinfo{person}{Luc Van~Gool}.} \bibinfo{year}{2022}\natexlab{b}.
\newblock \showarticletitle{MHFormer: Multi-Hypothesis Transformer for 3D Human Pose Estimation}. In \bibinfo{booktitle}{\emph{Proceedings of the IEEE/CVF Conference on Computer Vision and Pattern Recognition (CVPR)}}. \bibinfo{pages}{13147--13156}.
\newblock


\bibitem[Lin et~al\mbox{.}(2014)]%
        {cocodataset}
\bibfield{author}{\bibinfo{person}{Tsung-Yi Lin}, \bibinfo{person}{Michael Maire}, \bibinfo{person}{Serge Belongie}, \bibinfo{person}{James Hays}, \bibinfo{person}{Pietro Perona}, \bibinfo{person}{Deva Ramanan}, \bibinfo{person}{Piotr Doll{\'a}r}, {and} \bibinfo{person}{C~Lawrence Zitnick}.} \bibinfo{year}{2014}\natexlab{}.
\newblock \showarticletitle{Microsoft coco: Common objects in context}. In \bibinfo{booktitle}{\emph{Computer Vision--ECCV 2014: 13th European Conference, Zurich, Switzerland, September 6-12, 2014, Proceedings, Part V 13}}. Springer, \bibinfo{pages}{740--755}.
\newblock


\bibitem[Liu et~al\mbox{.}(2025)]%
        {liu2025tcpformer}
\bibfield{author}{\bibinfo{person}{Jiajie Liu}, \bibinfo{person}{Mengyuan Liu}, \bibinfo{person}{Hong Liu}, {and} \bibinfo{person}{Wenhao Li}.} \bibinfo{year}{2025}\natexlab{}.
\newblock \showarticletitle{TCPFormer: Learning Temporal Correlation with Implicit Pose Proxy for 3D Human Pose Estimation}.
\newblock \bibinfo{journal}{\emph{arXiv preprint arXiv:2501.01770}} (\bibinfo{year}{2025}).
\newblock


\bibitem[Mehta et~al\mbox{.}(2017)]%
        {mehta2017mpiinf}
\bibfield{author}{\bibinfo{person}{Dushyant Mehta}, \bibinfo{person}{Helge Rhodin}, \bibinfo{person}{Dan Casas}, \bibinfo{person}{Pascal Fua}, \bibinfo{person}{Oleksandr Sotnychenko}, \bibinfo{person}{Weipeng Xu}, {and} \bibinfo{person}{Christian Theobalt}.} \bibinfo{year}{2017}\natexlab{}.
\newblock \showarticletitle{Monocular 3d human pose estimation in the wild using improved cnn supervision}. In \bibinfo{booktitle}{\emph{2017 international conference on 3D vision (3DV)}}. IEEE, \bibinfo{pages}{506--516}.
\newblock


\bibitem[Menychtas et~al\mbox{.}(2023)]%
        {menychtas2023gait}
\bibfield{author}{\bibinfo{person}{Dimitrios Menychtas}, \bibinfo{person}{Nikolaos Petrou}, \bibinfo{person}{Ioannis Kansizoglou}, \bibinfo{person}{Erasmia Giannakou}, \bibinfo{person}{Athanasios Grekidis}, \bibinfo{person}{Antonios Gasteratos}, \bibinfo{person}{Vassilios Gourgoulis}, \bibinfo{person}{Eleni Douda}, \bibinfo{person}{Ilias Smilios}, \bibinfo{person}{Maria Michalopoulou}, {et~al\mbox{.}}} \bibinfo{year}{2023}\natexlab{}.
\newblock \showarticletitle{Gait analysis comparison between manual marking, 2D pose estimation algorithms, and 3D marker-based system}.
\newblock \bibinfo{journal}{\emph{Frontiers in Rehabilitation Sciences}}  \bibinfo{volume}{4} (\bibinfo{year}{2023}), \bibinfo{pages}{1238134}.
\newblock


\bibitem[Nibali et~al\mbox{.}(2021)]%
        {ASPset}
\bibfield{author}{\bibinfo{person}{Aiden Nibali}, \bibinfo{person}{Joshua Millward}, \bibinfo{person}{Zhen He}, {and} \bibinfo{person}{Stuart Morgan}.} \bibinfo{year}{2021}\natexlab{}.
\newblock \showarticletitle{{ASPset}: An Outdoor Sports Pose Video Dataset With {3D} Keypoint Annotations}.
\newblock \bibinfo{journal}{\emph{Image and Vision Computing}} (\bibinfo{year}{2021}), \bibinfo{pages}{104196}.
\newblock
\showISSN{0262-8856}
\href{https://doi.org/10.1016/j.imavis.2021.104196}{doi:\nolinkurl{10.1016/j.imavis.2021.104196}}


\bibitem[Pavllo et~al\mbox{.}(2019)]%
        {videopose3d}
\bibfield{author}{\bibinfo{person}{Dario Pavllo}, \bibinfo{person}{Christoph Feichtenhofer}, \bibinfo{person}{David Grangier}, {and} \bibinfo{person}{Michael Auli}.} \bibinfo{year}{2019}\natexlab{}.
\newblock \showarticletitle{3D human pose estimation in video with temporal convolutions and semi-supervised training}. In \bibinfo{booktitle}{\emph{Conference on Computer Vision and Pattern Recognition (CVPR)}}.
\newblock


\bibitem[Piergiovanni and Ryoo(2019)]%
        {piergiovanni2019early_mlbinjury}
\bibfield{author}{\bibinfo{person}{AJ Piergiovanni} {and} \bibinfo{person}{Michael~S Ryoo}.} \bibinfo{year}{2019}\natexlab{}.
\newblock \showarticletitle{Early detection of injuries in mlb pitchers from video}. In \bibinfo{booktitle}{\emph{Proceedings of the IEEE/CVF Conference on Computer Vision and Pattern Recognition Workshops}}. \bibinfo{pages}{0--0}.
\newblock


\bibitem[Shan et~al\mbox{.}(2023)]%
        {Shan_2023_ICCV_D3DP}
\bibfield{author}{\bibinfo{person}{Wenkang Shan}, \bibinfo{person}{Zhenhua Liu}, \bibinfo{person}{Xinfeng Zhang}, \bibinfo{person}{Zhao Wang}, \bibinfo{person}{Kai Han}, \bibinfo{person}{Shanshe Wang}, \bibinfo{person}{Siwei Ma}, {and} \bibinfo{person}{Wen Gao}.} \bibinfo{year}{2023}\natexlab{}.
\newblock \showarticletitle{Diffusion-Based 3D Human Pose Estimation with Multi-Hypothesis Aggregation}. In \bibinfo{booktitle}{\emph{Proceedings of the IEEE/CVF International Conference on Computer Vision (ICCV)}}. \bibinfo{pages}{14761--14771}.
\newblock


\bibitem[Sigal et~al\mbox{.}(2010)]%
        {Humaneva}
\bibfield{author}{\bibinfo{person}{Leonid Sigal}, \bibinfo{person}{Alexandru~O Balan}, {and} \bibinfo{person}{Michael~J Black}.} \bibinfo{year}{2010}\natexlab{}.
\newblock \showarticletitle{Humaneva: Synchronized video and motion capture dataset and baseline algorithm for evaluation of articulated human motion}.
\newblock \bibinfo{journal}{\emph{International journal of computer vision}} \bibinfo{volume}{87}, \bibinfo{number}{1} (\bibinfo{year}{2010}), \bibinfo{pages}{4--27}.
\newblock


\bibitem[Soroush~Mehraban(2024)]%
        {motionagformer2024}
\bibfield{author}{\bibinfo{person}{Babak~Taati Soroush~Mehraban, Vida~Adeli}.} \bibinfo{year}{2024}\natexlab{}.
\newblock \showarticletitle{MotionAGFormer: Enhancing 3D Human Pose Estimation with a Transformer-GCNFormer Network}. In \bibinfo{booktitle}{\emph{Proceedings of the IEEE/CVF Winter Conference on Applications of Computer Vision}}.
\newblock


\bibitem[Sun et~al\mbox{.}(2019)]%
        {sun2019deep_hrnet}
\bibfield{author}{\bibinfo{person}{Ke Sun}, \bibinfo{person}{Bin Xiao}, \bibinfo{person}{Dong Liu}, {and} \bibinfo{person}{Jingdong Wang}.} \bibinfo{year}{2019}\natexlab{}.
\newblock \showarticletitle{Deep high-resolution representation learning for human pose estimation}. In \bibinfo{booktitle}{\emph{Proceedings of the IEEE/CVF conference on computer vision and pattern recognition}}. \bibinfo{pages}{5693--5703}.
\newblock


\bibitem[Suzuki et~al\mbox{.}(2022)]%
        {suzuki2022automatic}
\bibfield{author}{\bibinfo{person}{Tomohiro Suzuki}, \bibinfo{person}{Kazuya Takeda}, {and} \bibinfo{person}{Keisuke Fujii}.} \bibinfo{year}{2022}\natexlab{}.
\newblock \showarticletitle{Automatic fault detection in race walking from a smartphone camera via fine-tuning pose estimation}. In \bibinfo{booktitle}{\emph{2022 IEEE 11th Global Conference on Consumer Electronics (GCCE)}}. IEEE, \bibinfo{pages}{631--632}.
\newblock


\bibitem[Suzuki et~al\mbox{.}(2024)]%
        {suzuki2024automatic}
\bibfield{author}{\bibinfo{person}{Tomohiro Suzuki}, \bibinfo{person}{Kazuya Takeda}, {and} \bibinfo{person}{Keisuke Fujii}.} \bibinfo{year}{2024}\natexlab{}.
\newblock \showarticletitle{Automatic Detection of Faults in Simulated Race Walking from a Fixed Smartphone Camera}.
\newblock \bibinfo{journal}{\emph{International Journal of Computer Science in Sport}} \bibinfo{volume}{23}, \bibinfo{number}{1} (\bibinfo{year}{2024}), \bibinfo{pages}{22--36}.
\newblock
\href{https://doi.org/doi:10.2478/ijcss-2024-0002}{doi:\nolinkurl{doi:10.2478/ijcss-2024-0002}}


\bibitem[Tanaka et~al\mbox{.}(2024)]%
        {MMSports2024_tanaka_fsjump3d}
\bibfield{author}{\bibinfo{person}{Ryota Tanaka}, \bibinfo{person}{Tomohiro Suzuki}, {and} \bibinfo{person}{Keisuke Fujii}.} \bibinfo{year}{2024}\natexlab{}.
\newblock \showarticletitle{3D Pose-Based Temporal Action Segmentation for Figure Skating: A Fine-Grained and Jump Procedure-Aware Annotation Approach}. In \bibinfo{booktitle}{\emph{Proceedings of the 7th ACM International Workshop on Multimedia Content Analysis in Sports}} (Melbourne VIC, Australia) \emph{(\bibinfo{series}{MMSports '24})}. \bibinfo{publisher}{Association for Computing Machinery}, \bibinfo{address}{New York, NY, USA}, \bibinfo{pages}{17–26}.
\newblock
\showISBNx{9798400711985}
\href{https://doi.org/10.1145/3689061.3689077}{doi:\nolinkurl{10.1145/3689061.3689077}}


\bibitem[Tanaka et~al\mbox{.}(2023a)]%
        {tanaka2023automatic_mmsports}
\bibfield{author}{\bibinfo{person}{Ryota Tanaka}, \bibinfo{person}{Tomohiro Suzuki}, \bibinfo{person}{Kazuya Takeda}, {and} \bibinfo{person}{Keisuke Fujii}.} \bibinfo{year}{2023}\natexlab{a}.
\newblock \showarticletitle{Automatic Edge Error Judgment in Figure Skating Using 3D Pose Estimation from a Monocular Camera and IMUs}. In \bibinfo{booktitle}{\emph{Proceedings of the 6th International Workshop on Multimedia Content Analysis in Sports}}. \bibinfo{pages}{41--48}.
\newblock


\bibitem[Tanaka et~al\mbox{.}(2023b)]%
        {tanaka2023automatic_gcce}
\bibfield{author}{\bibinfo{person}{Ryota Tanaka}, \bibinfo{person}{Tomohiro Suzuki}, \bibinfo{person}{Kazuya Takeda}, {and} \bibinfo{person}{Keisuke Fujii}.} \bibinfo{year}{2023}\natexlab{b}.
\newblock \showarticletitle{Automatic Edge Error Judgment in Figure Skating Using 3D Pose Estimation from Inertial Sensors}. In \bibinfo{booktitle}{\emph{2023 IEEE 12th Global Conference on Consumer Electronics (GCCE)}}. IEEE, \bibinfo{pages}{1099--1100}.
\newblock


\bibitem[Torvinen et~al\mbox{.}(2024)]%
        {torvinen2024evaluation_ski}
\bibfield{author}{\bibinfo{person}{Petra Torvinen}, \bibinfo{person}{Keijo~S Ruotsalainen}, \bibinfo{person}{Shuang Zhao}, \bibinfo{person}{Neil Cronin}, \bibinfo{person}{Olli Ohtonen}, {and} \bibinfo{person}{Vesa Linnamo}.} \bibinfo{year}{2024}\natexlab{}.
\newblock \showarticletitle{Evaluation of 3D markerless motion capture system accuracy during skate skiing on a treadmill}.
\newblock \bibinfo{journal}{\emph{Bioengineering}} \bibinfo{volume}{11}, \bibinfo{number}{2} (\bibinfo{year}{2024}), \bibinfo{pages}{136}.
\newblock


\bibitem[Trumble et~al\mbox{.}(2017)]%
        {TotalCapture}
\bibfield{author}{\bibinfo{person}{Matt Trumble}, \bibinfo{person}{Andrew Gilbert}, \bibinfo{person}{Charles Malleson}, \bibinfo{person}{Adrian Hilton}, {and} \bibinfo{person}{John Collomosse}.} \bibinfo{year}{2017}\natexlab{}.
\newblock \showarticletitle{Total Capture: 3D Human Pose Estimation Fusing Video and Inertial Sensors}. In \bibinfo{booktitle}{\emph{2017 British Machine Vision Conference (BMVC)}}.
\newblock


\bibitem[Uhlrich et~al\mbox{.}(2023)]%
        {uhlrich2023opencap}
\bibfield{author}{\bibinfo{person}{Scott~D Uhlrich}, \bibinfo{person}{Antoine Falisse}, \bibinfo{person}{{\L}ukasz Kidzi{\'n}ski}, \bibinfo{person}{Julie Muccini}, \bibinfo{person}{Michael Ko}, \bibinfo{person}{Akshay~S Chaudhari}, \bibinfo{person}{Jennifer~L Hicks}, {and} \bibinfo{person}{Scott~L Delp}.} \bibinfo{year}{2023}\natexlab{}.
\newblock \showarticletitle{OpenCap: Human movement dynamics from smartphone videos}.
\newblock \bibinfo{journal}{\emph{PLoS computational biology}} \bibinfo{volume}{19}, \bibinfo{number}{10} (\bibinfo{year}{2023}), \bibinfo{pages}{e1011462}.
\newblock


\bibitem[Von~Marcard et~al\mbox{.}(2018)]%
        {3dpw}
\bibfield{author}{\bibinfo{person}{Timo Von~Marcard}, \bibinfo{person}{Roberto Henschel}, \bibinfo{person}{Michael~J Black}, \bibinfo{person}{Bodo Rosenhahn}, {and} \bibinfo{person}{Gerard Pons-Moll}.} \bibinfo{year}{2018}\natexlab{}.
\newblock \showarticletitle{Recovering accurate 3d human pose in the wild using imus and a moving camera}. In \bibinfo{booktitle}{\emph{Proceedings of the European conference on computer vision (ECCV)}}. \bibinfo{pages}{601--617}.
\newblock


\bibitem[Washabaugh et~al\mbox{.}(2022)]%
        {washabaugh2022comparing_gait}
\bibfield{author}{\bibinfo{person}{Edward~P Washabaugh}, \bibinfo{person}{Thanikai~Adhithiyan Shanmugam}, \bibinfo{person}{Rajiv Ranganathan}, {and} \bibinfo{person}{Chandramouli Krishnan}.} \bibinfo{year}{2022}\natexlab{}.
\newblock \showarticletitle{Comparing the accuracy of open-source pose estimation methods for measuring gait kinematics}.
\newblock \bibinfo{journal}{\emph{Gait \& posture}}  \bibinfo{volume}{97} (\bibinfo{year}{2022}), \bibinfo{pages}{188--195}.
\newblock


\bibitem[Xu et~al\mbox{.}(2022)]%
        {xu2022vitpose}
\bibfield{author}{\bibinfo{person}{Yufei Xu}, \bibinfo{person}{Jing Zhang}, \bibinfo{person}{Qiming Zhang}, {and} \bibinfo{person}{Dacheng Tao}.} \bibinfo{year}{2022}\natexlab{}.
\newblock \showarticletitle{Vitpose: Simple vision transformer baselines for human pose estimation}.
\newblock \bibinfo{journal}{\emph{Advances in neural information processing systems}}  \bibinfo{volume}{35} (\bibinfo{year}{2022}), \bibinfo{pages}{38571--38584}.
\newblock


\bibitem[Yeung et~al\mbox{.}(2024)]%
        {yeung2024autosoccerpose}
\bibfield{author}{\bibinfo{person}{Calvin Yeung}, \bibinfo{person}{Kenjiro Ide}, {and} \bibinfo{person}{Keisuke Fujii}.} \bibinfo{year}{2024}\natexlab{}.
\newblock \showarticletitle{Autosoccerpose: Automated 3d posture analysis of soccer shot movements}. In \bibinfo{booktitle}{\emph{Proceedings of the IEEE/CVF Conference on Computer Vision and Pattern Recognition}}. \bibinfo{pages}{3214--3224}.
\newblock


\bibitem[Yeung et~al\mbox{.}(2025)]%
        {yeung2025athletepose3d}
\bibfield{author}{\bibinfo{person}{Calvin Yeung}, \bibinfo{person}{Tomohiro Suzuki}, \bibinfo{person}{Ryota Tanaka}, \bibinfo{person}{Zhuoer Yin}, {and} \bibinfo{person}{Keisuke Fujii}.} \bibinfo{year}{2025}\natexlab{}.
\newblock \showarticletitle{AthletePose3D: A benchmark dataset for 3D human pose estimation and kinematic validation in athletic movements}. In \bibinfo{booktitle}{\emph{Proceedings of the Computer Vision and Pattern Recognition Conference}}. \bibinfo{pages}{5945--5956}.
\newblock


\bibitem[Zhang et~al\mbox{.}(2022)]%
        {mixste}
\bibfield{author}{\bibinfo{person}{Jinlu Zhang}, \bibinfo{person}{Zhigang Tu}, \bibinfo{person}{Jianyu Yang}, \bibinfo{person}{Yujin Chen}, {and} \bibinfo{person}{Junsong Yuan}.} \bibinfo{year}{2022}\natexlab{}.
\newblock \showarticletitle{MixSTE: Seq2seq Mixed Spatio-Temporal Encoder for 3D Human Pose Estimation in Video}. In \bibinfo{booktitle}{\emph{Proceedings of the IEEE/CVF Conference on Computer Vision and Pattern Recognition (CVPR)}}. \bibinfo{pages}{13232--13242}.
\newblock


\bibitem[Zhu et~al\mbox{.}(2023)]%
        {motionbert2022}
\bibfield{author}{\bibinfo{person}{Wentao Zhu}, \bibinfo{person}{Xiaoxuan Ma}, \bibinfo{person}{Zhaoyang Liu}, \bibinfo{person}{Libin Liu}, \bibinfo{person}{Wayne Wu}, {and} \bibinfo{person}{Yizhou Wang}.} \bibinfo{year}{2023}\natexlab{}.
\newblock \showarticletitle{MotionBERT: A Unified Perspective on Learning Human Motion Representations}. In \bibinfo{booktitle}{\emph{Proceedings of the IEEE/CVF International Conference on Computer Vision}}.
\newblock


\end{thebibliography}










\end{document}